\pdfoutput=1

\documentclass[11pt]{article}

\usepackage{ACL2023}

\usepackage{times}
\usepackage{latexsym}
\usepackage{booktabs}
\usepackage{multirow}

\usepackage{graphicx}

\usepackage[T1]{fontenc}

\usepackage[utf8]{inputenc}

\usepackage{microtype}

\usepackage{inconsolata}
\usepackage{amssymb}
\usepackage{amsmath}
\usepackage{listings}
\usepackage{bbding} %
\usepackage{fdsymbol} %
\usepackage{url}

\definecolor{codegreen}{rgb}{0,0.6,0}
\definecolor{codegray}{rgb}{0.5,0.5,0.5}

\definecolor{backcolour}{RGB}{245,248,250}
\definecolor{emph}{RGB}{166,88,53}
\definecolor{nightblue}{RGB}{9,49,105}
\definecolor{keywords}{RGB}{207,33,46}
\definecolor{lightpurple}{RGB}{130,81,223}

\lstdefinestyle{mystyle}{
    backgroundcolor=\color{backcolour},   
    commentstyle=\color{codegreen},
    keywordstyle=\color{keywords},
    stringstyle=\color{nightblue},
    basicstyle=\fontsize{7}{8}\ttfamily,
    breakatwhitespace=true,         
    breaklines=true,                 
    captionpos=b,                    
    keepspaces=true,                 
    numberstyle=\tiny\color{codegray},
    numbersep=2pt,                  
    showspaces=false,                
    showstringspaces=false,
    showtabs=false,                  
    tabsize=2,
    emph={dspy},
    emphstyle={\color{lightpurple}},
    linewidth=1\columnwidth,
    frame=tb,    
    xrightmargin=0pt,
    xleftmargin=0.23cm,
    numbers=left,
    aboveskip=0.2cm,
    belowskip=0.1cm,
}

\newcommand{\irera}{\texttt{Infer--Retrieve--Rank}}
\newcommand{\ire}{\texttt{Infer--Retrieve}}

\title{In-Context Learning for Extreme Multi-Label Classification}

\author{First Author \\
  Affiliation / Address line 1 \\
  Affiliation / Address line 2 \\
  Affiliation / Address line 3 \\
  \texttt{email@domain} \\\And
  Second Author \\
  Affiliation / Address line 1 \\
  Affiliation / Address line 2 \\
  Affiliation / Address line 3 \\
  \texttt{email@domain} \\}

\author{Karel D'Oosterlinck$^{1,2,*}$, Omar Khattab$^2$, Fran\c{c}ois Remy$^1$, \\  \textbf{Thomas Demeester$^1$, Chris Develder$^1$, Christopher Potts$^2$} \\ $^1$Ghent University -- imec\qquad$^2$Stanford University\\
$^*$\texttt{karel.doosterlinck@ugent.be}
}

\begin{document}
\maketitle

\begin{abstract}

Multi-label classification problems with thousands of classes are hard to solve with in-context learning alone, as language models (LMs) might lack prior knowledge about the precise classes or how to assign them, and it is generally infeasible to demonstrate every class in a prompt. We propose a general program, \irera{}, that defines multi-step interactions between LMs and retrievers to efficiently tackle such problems. 
We implement this program using the \texttt{DSPy} programming model, which specifies in-context systems in a declarative manner, and use \texttt{DSPy} optimizers to tune it towards specific datasets by bootstrapping only tens of few-shot examples.
Our primary extreme classification program, optimized separately for each task, attains state-of-the-art results across three benchmarks (HOUSE, TECH, TECHWOLF). We apply the same program to a benchmark with vastly different characteristics and attain competitive performance as well (BioDEX).
Unlike prior work, our proposed solution requires no finetuning, is easily applicable to new tasks, alleviates prompt engineering, and requires only tens of labeled examples. 
Our code is public at \url{https://github.com/KarelDO/xmc.dspy}.

\end{abstract}

\section{Introduction}

Extreme multi-label classification (XMC) tasks are hard to solve with in-context learning alone. Language models (LMs) might lack prior knowledge about the precise classes, and the sheer number of available classes---often upwards of 10,000---generally means it is infeasible even to demonstrate every class in a prompt.
To deal with this, some recent efforts make multiple LM calls at inference time~\citep{zhu2023icxml}, while others prompt LMs to generate synthetic data for finetuning~\citep{decorte2023extreme, clavie2023large}. These methods can be configured to work well, but they all have manual ``knobs'' like prompts and other hyperparameters that make applying them to new datasets, metrics, or LMs challenging. 

\begin{figure*}[]
    \centering
    \includegraphics[width=0.8\linewidth]{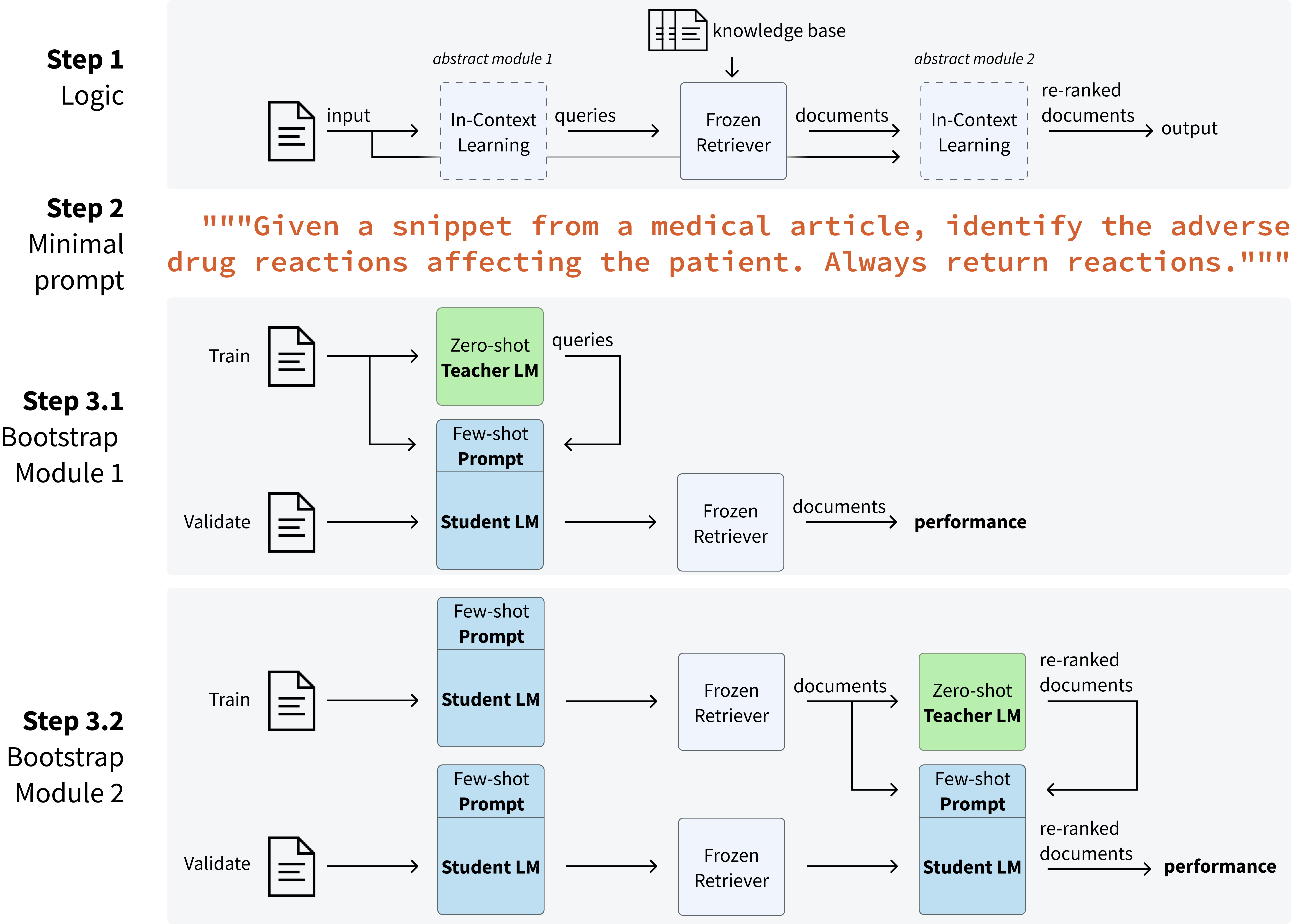}
    \caption{We propose \texttt{Infer-Retrieve-Rank}, an efficient in-context learning program for multi-label classification with an extreme amount of classes ($\geq$ 10,000). Given an input, a first in-context learning module predicts queries which route to a frozen retriever. The retrieved documents are re-ranked by a second in-context module (\textbf{Step~1}). 
    Given a minimal prompt (\textbf{Step~2}),
    a zero-shot \textbf{Teacher LM} bootstraps  demonstrations to optimize the few-shot \textbf{Student LM} (\textbf{Step~3}). 
    Optimization using $\approx$50 labeled inputs can yield state-of-the-art results, using only $\approx$20 Teacher and $\approx$1,500 Student calls. The (optimization) logic is expressed using the \texttt{DSPy} programming model.
    }
    \label{fig:schematic}
\end{figure*}

In this paper, we show that simple programs written using the \texttt{DSPy} programming model \citep{khattab2023dspy}  support powerful, highly general approaches to XMC tasks. 
\texttt{DSPy} allows us to separately specify the modular \emph{program} of our method and how it should be \emph{optimized} towards different datasets. We propose a simple in-context program for XMC tasks called \irera{} (\texttt{IReRa}, Figure~\ref{fig:schematic}, Step 1). First, an LM processes the input document and guesses a set of applicable terms (\texttt{Infer}). Then, a retriever relates each predicted term to the actual label space (\texttt{Retrieve}). Finally, an LM is used to rerank retrieved labels (\texttt{Rank}). Crucially, we use a frozen retriever and frozen LMs. The key insights of \irera{} is that such a frozen retriever can be made much more flexible if the LM learns in-context how to predict relevant queries and interpret the retrieved results.

The underlying LMs, retriever, and prompts are considered \emph{hyperparameters} of the \texttt{IReRa} program, which can be tuned automatically or easily configured. Using only 10 unlabeled training inputs, and $\approx$50 labeled validation examples, we bootstrap a few-shot prompt for our two LM components, using a zero-shot teacher LM with a minimal seed-prompt (Figure~\ref{fig:schematic}, Step 2). \texttt{DSPy}'s compilation abstraction handles this nicely; it takes the program logic we've already defined, instantiates it with a teacher LM, processes the unlabeled training examples, generates zero-shot labels for each of the program steps, and picks the best labels to put in a few-shot prompt based on validation performance. Because our program consists of two in-context modules, we propose to bootstrap them sequentially (Figure~\ref{fig:schematic}, Step 3).

In our experiments, we instantiate the \texttt{Infer} module with a  \texttt{Llama-2-7b-chat} model \citep{touvron2023llama}, while the teacher model used for bootstrapping is \texttt{GPT-3.5}. The \texttt{Rank} module is instantiated and bootstrapped both by a \texttt{GPT-4} model.

Adapting \irera{} to a new dataset can be as simple as (i) writing a new minimal zero-shot prompt, (ii) configuring which LMs to use, and (iii) running the optimization procedure. We optimize this program separately towards 4 XMC datasets: one dataset involving the extracting and coding of adverse drug events expressed in biomedical literature (BioDEX; \citealt{d2023biodex}) and three datasets involving the labeling of job vacancy snippets with the required competencies they express (HOUSE, TECH, TECHWOLF; \citealt{zhang-etal-2022-skillspan, decorte2022design, decorte2023extreme}). Our program attains state-of-the-art results on the job vacancy datasets, and gets meaningful traction on the harder biomedical task---without finetuning, without prompt engineering, and by using only $\approx$50 labeled examples. We find that the optimization is a consistent driver of performance across tasks.

\section{Related Work}

The canonical way of tackling an extreme classification problem involves either finetuning a specialized retriever over the label space or finetuning one binary classifier per class~\citep{decorte2022design, decorte2023extreme, clavie2023large}. These methods require a lot of data, since every one of the many classes requires at least a few labeled examples. To avoid manual data labeling, researchers use distant supervision~\citep{decorte2022design}, bootstrap synthetic data using LLMs%
~\citep{decorte2023extreme, clavie2023large, de-raedt-etal-2023-idas}, or finetune retrievers on adjacent problems where data is available~\citep{remy-etal-2022-biolord}. At inference time, an additional LLM call can be used to rerank a list of generated candidate labels to further increase performance \citep{clavie2023large}.

\citet{zhu2023icxml} use multiple \texttt{GPT-3.5} calls combined with retrieval at inference-time to bootstrap a synthetic prompt per input, infer labels, and rerank them. While they do not use any finetuning, they require many LLM and retrieval calls per input. They evaluate on two recommendation tasks, where inputs and outputs are the same type of documents. \citet{Bhatia16} formulate many recommendation tasks under the XMC setting. Instead, we consider XMC tasks where inputs and outputs are not of similar shape, and more inference or information extraction is needed.

Our \irera{} program does not rely on finetuning or many LLM calls per input, making it efficient to develop and deploy. \irera{} can achieve state-of-the-art performance using only $\approx$50 labeled examples. Unlike prior work, our program logic is defined in a modular and declarative manner, and can be seamlessly applied to different benchmarks given a minimal seed-prompt. Optimization happens automatically and can resolve in as little as ten minutes. The choice of LMs and retrievers can be configured, ensuring relevance when stronger components become available. Finally, we write at most one seed-prompt per task and in-context module, and let optimization---not iterative prompt engineering---take care of increasing performance.

\section{\irera{}}

The program for \irera{} is given in Code Snippet~\ref{lst:code}, with minor alterations for brevity. First, an LM is used to predict queries given the input (\texttt{Infer}). 
The retriever outputs a ranking over all labels based on maximum cosine embedding similarity with the queries (\texttt{Retrieve}). The top labels are reranked by another LM (\texttt{Rank}).

\renewcommand{\lstlistingname}{Code Snippet} %
\begin{lstlisting}[caption={\texttt{DSPy} code for \irera{} with minor alterations for brevity.}, label={lst:code}, language=Python,breaklines=true,showstringspaces=false,literate={í}{{\'i}}1]
class InferRetrieveRank(dspy.Module):
  def __init__(self, infer_sig, rank_sig, retr):
    # Initialize LM modules with Signatures
    self.infer = dspy.ChainOfThought(infer_sig)
    self.rank = dspy.ChainOfThrought(rank_sig)
    self.retrieve = retr
  
  def forward(self, text: str) -> Prediction:
    # Predict with LM
    preds = self.infer(text).completions.labels
  
    # Parse LM output
    preds = extract_labels_from_strings(preds)

    # Use LM outputs to retrieve labels
    labels = self.retrieve(preds)

    # Use LM to rerank labels
    labels = self.rank(text, labels)

    return dspy.Prediction(labels=labels)
\end{lstlisting}

Not all labels occur with equal frequency. Finetuned methods can implicitly learn this bias given enough data. If available, we propose to use the prior probability $p_i$ for the $i$-th label to reweigh the retrieval similarity $s_i$ to account for this. The updated scores $\tilde{s}_i$ as defined below are the output of the \texttt{Retrieve} module. $A$ is a scalar hyperparameter controlling the strength of the prior update:
\begin{equation*}
    \tilde{s}_i = s_i \cdot \log_{10}(A \cdot p_i + 10)
\end{equation*}

\section{Seed-prompts}\label{sec:appendix_prompts}
To apply \irera{} to a dataset, a minimal seed-prompt needs to define the behavior of each in-context module. Code Snippet~\ref{lst:code_signature} contains the prompt for the \texttt{Infer} module on the BioDEX dataset, neatly organized using the \texttt{DSPy} \texttt{Signature} abstraction. This seed-prompt defines a task description in the docstring, and input and output fields with descriptions and formatting information. The \texttt{Signature} serves as skeleton for both zero- and few-shot prompts.

\renewcommand{\lstlistingname}{Code Snippet} %
\begin{lstlisting}[caption={\texttt{DSPy} \texttt{Signature} for BioDEX \texttt{Infer}.}, label={lst:code_signature}, language=Python,breaklines=true,showstringspaces=false,literate={í}{{\'i}}1]
class BiodexInferSignature(dspy.Signature):
    """Given a snippet from a medical article, identify the adverse drug reactions affecting the patient. Always return reactions."""

    text = dspy.InputField(prefix="Article:")
    output = dspy.OutputField(
        prefix="Reactions:",
        desc="list of comma-separated adverse drug reactions"
    )
\end{lstlisting}

The prompt for the BioDEX \texttt{Rank} module is given in Code Snippet~\ref{lst:code_signature_rank_biodex}. We use the same prompts for all three job vacancy datasets, they are given in given in Code Snippets~\ref{lst:code_signature_infer_esco} and~\ref{lst:code_signature_rank_esco} for the \texttt{Infer} and \texttt{Rank} modules respectively. 
Note how the prompts share most of their content: adapting \irera{} can be as easy as concisely describing the input and output fields.

\renewcommand{\lstlistingname}{Code Snippet} %
\begin{lstlisting}[caption={\texttt{DSPy} \texttt{Signature} for BioDEX \texttt{Rank}.}, label={lst:code_signature_rank_biodex}, language=Python,breaklines=true,showstringspaces=false,literate={í}{{\'i}}1]
class BiodexRankSignature(dspy.Signature):
    """Given a snippet from a medical article, pick the 10 most applicable adverse reactions from the options that are directly expressed in the snippet."""

    text = dspy.InputField(prefix="Article:")
    options = dspy.InputField(
        prefix="Options:",
        desc="List of comma-separated options to choose from"
    )
    output = dspy.OutputField(
        prefix="Reactions:",
        desc="list of comma-separated adverse drug reactions"
    )
\end{lstlisting}

\renewcommand{\lstlistingname}{Code Snippet} %
\begin{lstlisting}[caption={\texttt{DSPy} \texttt{Signature} for ESCO \texttt{Infer}.}, label={lst:code_signature_infer_esco}, language=Python,breaklines=true,showstringspaces=false,literate={í}{{\'i}}1]
class EscoInferSignature(dspy.Signature):
    """Given a snippet from a job vacancy, identify all the ESCO job skills mentioned. Always return skills."""

    text = dspy.InputField(prefix="Vacancy:")
    options = dspy.InputField(
        prefix="Options:",
        desc="List of comma-separated options to choose from"
    )
    output = dspy.OutputField(
        prefix="Skills:",
        desc="list of comma-separated ESCO skills"
    )
\end{lstlisting}

\renewcommand{\lstlistingname}{Code Snippet} %
\begin{lstlisting}[caption={\texttt{DSPy} \texttt{Signature} for ESCO \texttt{Rank}.}, label={lst:code_signature_rank_esco}, language=Python,breaklines=true,showstringspaces=false,literate={í}{{\'i}}1]
class EscoRankSignature(dspy.Signature):
    """Given a snippet from a job vacancy, pick the 10 most applicable skills from the options that are directly expressed in the snippet."""

    text = dspy.InputField(prefix="Vacancy:")
    options = dspy.InputField(
        prefix="Options:",
        desc="List of comma-separated options to choose from"
    )
    output = dspy.OutputField(
        prefix="Skills:",
        desc="list of comma-separated ESCO skills"
    )
\end{lstlisting}

\section{Metrics}

We measure the rank-precision ($\textit{RP}$) of the produced rankings, which is the precision of the ranking at the rank equal to the number of total gold labels. Specifically we consider the rank-precision at $K$ ($\textit{RP}@K$; defined below). Given a gold number of labels $R_n$ for input $n$, the $\textit{RP}@K$ measures $\text{precision}@K$ when $K \leq R_n$ and $\text{recall}@K$ when $K \geq R_n$.%
\footnote{When $K = R_n$, the precision and the recall of the ranking are by definition equal \citep{aslam2005geometric}.}
$\textit{Rel}(n,k) = 1$ if the $k$-th output for input $n$ in the ranking was relevant, else $0$.

\vspace{-1em}
\begin{equation*}
    \textit{RP@K} = \frac{1}{N} \sum\limits_{n=1}^{N} \frac{1}{\min(K, R_n)}\sum\limits_{k=1}^{K} \textit{Rel}(n,k)
\end{equation*}

\section{Data} \label{sec:data}
We evaluate our method and baselines on four extreme classification datasets, one in the biomedical field and three in the field of human-resources.

\paragraph{BioDEX:} The BioDEX dataset (Biomedical Drug Event eXtraction;~\citealt{d2023biodex}) consists of biomedical papers containing various descriptions of adverse drug events and associated expert-created labels for the exact type of medical reaction discussed. These events are encoded in the MedDRA ontology (Medical Dictionary for Regulatory Activities;~\citealt{brown1999medical}), a set of $\approx$24,300 standardized medical reaction. Inputs can be very long (half of inputs have upwards of $\approx$20,000 characters), and biomedical domain knowledge is needed to infer the correct reactions (not all medical reactions need to be reported, only the \emph{adverse} ones). BioDEX models a crucial step in real-world drug safety pipelines. We use a subset of 10 training, 50 validation, and 250 test examples for our experiments. The median amount of labels per input is 3 while the 95th percentile is 14.

\paragraph{ESCO:} The ESCO ontology \citep{ESCO2017} contains $\approx$13,900 distinct concepts used to encode skills, competences, qualifications, and occupations. We consider three datasets each containing snippets (typically one sentence) of online job vacancies in English with their relevant ESCO labels. We use the HOUSE, TECH, and TECHWOLF datasets~\citep{zhang-etal-2022-skillspan, decorte2022design, decorte2023extreme}. We take 10 examples each from the HOUSE and TECH validation sets as training examples, and keep the remaining 51 and 65 examples as validation respectively. TECHWOLF has no validation or training split, so we use the train and validation split of HOUSE instead. HOUSE, TECH, and TECHWOLF respectively contain 262, 338, and 326 test examples. The median amount of labels per input across these datasets is 1 and the 95th percentile is 4.

\section{Experiments and Results}

Table~\ref{table:results_test} gives test results for all models and tasks.

\begin{table*}[tp]

\renewcommand{\Checkmark}{Yes}
\renewcommand{\XSolidBrush}{No}

\small
\centering
\begin{tabular}{@{}l cc cc cc cc cc}
\toprule
                                                                                                                                                                                                                         & \multicolumn{2}{c}{HOUSE} & \multicolumn{2}{c}{TECH} & \multicolumn{2}{c}{TECHWOLF} 
                                                                                                                                                                                                                         & \multicolumn{2}{c}{BioDEX}
                                                                                                                            & 
                                                                                                                                      \\

\cmidrule(lr){2-3}
\cmidrule(lr){4-5}
\cmidrule(lr){6-7}
\cmidrule(lr){8-9}

                                                                                                                                      & $\textit{RP}5$                                                  & $\textit{RP}10$                                                 & $\textit{RP}5$         & $\textit{RP}10$       & $\textit{RP}5$        & $\textit{RP}10$       & $\textit{RP}5$         & $\textit{RP}10$ & &          \\ \midrule
\textbf{Baselines} \\ \midrule
\texttt{prior}                                                                                                                                                               & \phantom{0}2.90          & \phantom{0}2.97        & \phantom{0}1.63        & \phantom{0}1.63        & \phantom{0}0.00           & \phantom{0}2.57 & 20.42                                        & 21.51          \\
\texttt{exact-match}                                                                                                                                                                            & \phantom{0}5.89         & \phantom{0}5.89        & \phantom{0}4.09        & \phantom{0}4.09        & \phantom{0}3.43          & \phantom{0}3.43  & \phantom{0}9.60                                                   & 15.16        \\
\texttt{naive-retrieve}                                                            & 26.17        & 36.76       & 39.60        & 49.79       & 33.48         & 42.13 & 10.99 & 11.71        \\  \midrule%
\textbf{Programs} &\multicolumn{8}{l}{ (each program requires 10 training and $\approx$50 validation examples)}          & Finetune & \#~LM calls \\ \midrule
\irera{}                                                                                                                                                               & \textbf{56.50}        &  \underline{65.76}      &         \underline{59.61}        & \textbf{70.23} & \textbf{57.04}         &\textbf{65.17}  & 24.73                                                  & 27.67 & \XSolidBrush & $\approx$1,520        \\ 
\quad$-$ optimize \texttt{Rank}                                                                                                                                                               & \underline{52.19}        & \textbf{66.51}        & 56.77        & \textbf{70.58}        & 51.34          & \underline{62.32} & 24.59                                                 & 28.55& \XSolidBrush & $\approx$1,010        \\ 

\ire{}                                                                                                                                                              & 42.47        & 52.62       & 55.01       & 62.45       & 47.49         & 56.50 & 20.69                                                 & 24.77& \XSolidBrush & $\approx$1,010        \\ 
\quad$-$ optimize \texttt{Infer}                                                                                                                                                                      & 20.23        & 30.69       & 21.76       & 33.42       & 22.15         & 29.69 & 15.40                                                & 15.76& \XSolidBrush & $0$         \\ \midrule
\textbf{Finetuned systems} & & & & & & & & &   & \#~Train size  \\ \midrule
\texttt{retrieve} \textsuperscript{$\clubsuit$}                                                                                                                                          & 45.74        & 55.95           & 54.62       & 66.24           & \underline{54.57}         & \underline{62.55} & /                                                     & /& \Checkmark & $\approx$138,000              \\
\texttt{retrieve-rankGPT3.5} \textsuperscript{$\diamondsuit$}                                                     & 43.57        & 51.44       & 52.50       & 59.75       & /             & / & /                                                     & /& \Checkmark & $\approx$555,000  \\       
\texttt{retrieve-rankGPT4} \textsuperscript{$\diamondsuit$}                                                      & \textbf{56.67}        & 61.02       & \textbf{61.50}       & \underline{68.94}       & /             & / & /                                                     & /& \Checkmark & $\approx$555,000   \\  
\texttt{seq2seq-prior} \textsuperscript{$\heartsuit$}
  & / & / & / & / & / & / & \underline{33.78} & \underline{35.52}& \Checkmark & $\approx$11,500\\
\texttt{10$\times$seq2seq-prior} \textsuperscript{$\heartsuit$}
 & / & / & / & / & / & / & \textbf{42.94} & \textbf{46.84}& \Checkmark & $\approx$11,500\\ \bottomrule      
\end{tabular}
\caption{Test results for baselines, programs, and finetuned systems on the HOUSE, TECH, TECHWOLF, and BioDEX extreme multi-label classification tasks. 
Metrics are rank-precision ($\textit{RP}$) at 5 and at 10. 
Our instantiation of \irera{} uses a \texttt{Llama-2-7b-chat} model to \texttt{Infer}, a frozen \texttt{BioLORD} or \texttt{all-mpnet-base-v2} to \texttt{Retrieve}, and a \texttt{GPT-4} model to \texttt{Rank}.
\irera{} can attain state-of-the-art results compared to specialized systems while requiring no finetuning and multiple orders of magnitude less data. 
Each program requires an amount of LM calls to bootstrap, which is compared with the training size used by finetuned systems.
Best results within a 0.5 interval in \textbf{bold}, second best results \underline{underlined}. 
The finetuned system results are taken from \textsuperscript{$\clubsuit$}\citet{decorte2023extreme} and  \textsuperscript{$\diamondsuit$}\citet{clavie2023large} where available, or adapted from \textsuperscript{$\heartsuit$}\citet{d2023biodex}.
}
\label{table:results_test}
\end{table*}

\paragraph{Baselines}

We evaluate a set of baselines across the four tasks. First, we evaluate a ranking equal to the prior statistic over all the labels (\texttt{prior}). For BioDEX, we estimate these priors across all the BioDEX training data. For the ESCO datasets, we use the priors distributed by~\citet{decorte2023extreme}, which are calculated from a private training set. Subsequently we evaluate the performance of exactly matching the label names in the input document (\texttt{exact-match}). Finally, we embed the input document with an off-the-shelf retriever and retrieve over label embeddings. We use the pre-trained \texttt{all-mpnet-base-v2} model~\citep{reimers-gurevych-2019-sentence} for ESCO-tasks and \texttt{BioLORD}~\citep{remy-etal-2022-biolord}, a biomedical retriever, for BioDEX (\texttt{naive-retrieve}).

Through these baselines, an interesting distinction between BioDEX and the ESCO-tasks emerges. Off-the-shelf retrieval is much stronger on ESCO-tasks. 
We hypothesize this is due to the shape of the input documents. Entire biomedical publications are hard to compress into a single vector---especially with an off-the-shelf retriever. The short vacancy snippets are easier to handle.

\paragraph{\irera{}}

We instantiate the \texttt{Infer} module with a \texttt{Llama-2-7b-chat} student LM and \texttt{GPT-3.5-turbo} teacher LM. The \texttt{Rank} module uses \texttt{GPT-4} as both student and teacher. 
The seed-prompts are given in Code Snippets~\ref{lst:code_signature},~\ref{lst:code_signature_rank_biodex},~\ref{lst:code_signature_infer_esco}, and ~\ref{lst:code_signature_rank_esco}.

We optimize \irera{} for $\textit{RP}@10$ performance on each dataset separately. Each run involves 10 unlabeled training examples and $\approx$50 labeled validation examples. Every run incurs $\approx$20 teacher model calls and $\approx$1,500 student model calls, and can complete in tens of minutes. We use \texttt{dspy}'s \texttt{BootstrapFewShotWithRandomSearch} class to automate the prompt bootstrapping procedure. A detailed breakdown of optimization and inference costs, in function of the different LMs used, is given in Section~\ref{sec:appendix_cost}. We set the prior hyperparameter $A$ to 0 for ESCO-tasks and 1000 for BioDEX, based on a handful of validation runs.

For ESCO-tasks, we compare with the best finetuned systems from the literature. \texttt{retrieve} denotes the retriever of~\citet{decorte2023extreme}, \texttt{retrieve-rankGPT3.5/4} denotes the system with inference-time reranking of~\citet{clavie2023large}. For BioDEX, we slightly alter the method of~\citet{d2023biodex}: we take a \texttt{FLAN-T5-Large} model \citep{chung2022scaling} and train it to output a comma-separated list of reaction labels given a chunk of the input paper (the original BioDEX system was trained to output many attributes, of which medical reactions was only one). This model does not directly produce a ranking, so if a reaction is not predicted we add it in order of the prior (\texttt{seq2seq-prior}). We also consider sampling 10 generations from the model and majority voting the reactions (\texttt{10$\times$seq2seq-prior}).

\irera{} achieves state-of-the-art performance across all ESCO-tasks. Through a set of ablations, we find that each optimization step and module improves performance. Notable, the \texttt{Infer-Retrieve} system, which ablates the \texttt{Rank} module, can still attain competitive results despite using only one open-source LM and frozen retriever. \irera{} does not beat our finetuned system on BioDEX, but adding a \texttt{Rank} module or optimizing \texttt{Infer} consistently improves performance, indicating that programs can support a general approaches to extreme multi-label classification across a variety of datasets with different characteristics.

\section{Program Cost Breakdown}
\label{sec:appendix_cost}

\begin{table*}[]
\begin{tabular}{@{}lll cc c@{}}
\toprule
                    & \multicolumn{2}{l}{Configuration}                    & \multicolumn{2}{c}{Optimize Calls}                   & Calls / Input                  \\

\cmidrule(lr){2-3}
\cmidrule(lr){4-5}
\cmidrule(lr){6-6}
                    & LM      & Teacher         & LM                       & Teacher        & LM                            \\ \midrule

\textbf{Modules}      &                  &                    &          &                                                             \\ \midrule
\texttt{Infer}               & \texttt{Llama}            & \texttt{GPT3.5}             & $\approx$500                         & $\approx$10          & 1                                               \\
\texttt{Retrieve}            & \texttt{mpnet} (or similar)           & \texttt{None}                  & 0                         & 0                 & 1                                               \\
\texttt{Rank}                & \texttt{GPT4}             & \texttt{GPT4}               & $\approx$500                         & $\approx$10          & 1                                               \\ \midrule
\textbf{Programs}     &   &  & & &  \\ \midrule
\irera{} & \texttt{\texttt{Llama}}-\texttt{mpnet}-\texttt{GPT4} & \texttt{GPT3.5}-\texttt{None}-\texttt{GPT4}   & $\approx$1,500                       & $\approx$20          & 3                                               \\
\quad$-$ optimize \texttt{Rank}       & \texttt{\texttt{Llama}}-\texttt{mpnet}-\texttt{GPT4} & \texttt{GPT3.5}-\texttt{None}        & $\approx$1,000                       & $\approx$10          & 3                                               \\
\texttt{Infer-Retrieve}      & \texttt{\texttt{Llama}}-\texttt{mpnet}      & \texttt{GPT3.5}-\texttt{None}        & $\approx$1,000                       & $\approx$10          & 2                                               \\
\quad$-$ optimize \texttt{Infer}      & \texttt{\texttt{Llama}}-\texttt{mpnet}      & \texttt{None}-\texttt{None}               & 0                         & 0                 & 2                                               \\ \bottomrule
\end{tabular}
\caption{Breakdown of configuration and costs associated with our \textbf{Modules} and \textbf{Programs}. A module is configured with a single LM and Teacher. A program inherits all LMs of its modules, and all Teachers of its \emph{optimized} modules. Per module, the optimization procedure requires $O(\textit{train})$ Teacher calls and $\textit{num\_programs} \cdot O(\textit{val})$ LM calls, where $\textit{num\_programs}$ is a hyperparameter controlling how many programs to try during optimization. During inference, 1 LM call is used per module. Programs inherit the optimization calls of their \emph{optimized} modules, and the inference calls of all modules. While \texttt{Retrieve} is not directly optimized, it does contribute inference calls to the optimization of \texttt{Infer-Retrieve} because the \texttt{Retrieve} module is in the optimization loop. In this work, we use 10 training examples, $\approx$50 validation examples, and set $\textit{num\_programs}$ to 10.}
\label{table:cost}
\end{table*}

Table~\ref{table:cost} outlines the optimization and inference calls associated with our modules and programs. A module is always instantiated with one LM, but can also have another Teacher LM if it is optimizable. In our case, the \texttt{Infer} and \texttt{Rank} modules have a Teacher while \texttt{Retrieve} does not. Optimizing a module given its inputs incurs $O(\textit{train})$ calls from the Teacher model and $\textit{num\_programs} \cdot O(\textit{validation})$ calls from the Student LM, where $\textit{train}$ and $\textit{validation}$ denote the sizes of the training and validation sets respectively, and $\textit{num\_programs}$ controls how many different bootstrapped prompts to try in the optimization process. In our work, this results in $\approx$10 Teacher and $\approx$500 LM calls per module.

Programs inherit the configuration and calls from their constituent modules. Teacher optimization calls are only inherited if the module is actually optimized, LM optimization calls are inherited if the module is in the loop for another optimized module, and inference calls are always inherited.

Table~\ref{table:cost} makes it easy to express the cost associated with any program. For example, our state-of-the-art \irera{} program requires approximately 500 \texttt{Llama}, 500 \texttt{mpnet}, 10 \texttt{GPT3.5}, and 510 \texttt{GPT4} calls to optimize. This is calculated as follows. First, 500 \texttt{Llama} and 500 \texttt{mpnet} student LM calls and 10 \texttt{GPT3.5} teacher calls are needed to optimize \texttt{Infer-Retrieve}. Then an additional 500 \texttt{GPT4} student LM calls and 10 \texttt{GPT4} teacher calls are needed to optimize \texttt{Rank}. Notice how \texttt{Infer-Retrieve} is in the loop while \irera{} is optimized. Because of the left-to-right nature of this optimization procedure, we can cache the inference calls of \texttt{Infer-Retrieve}, saving us the cost of executing them again when optimizing \texttt{Rank}. Per new input, the program incurs 1 call for each \texttt{Llama}, \texttt{mpnet}, and \texttt{GPT4} LM.   

The finetuned systems we compare to in Table~\ref{table:results_test} all have a much higher start-up cost, in part due to the need for labeled finetuning data---which some systems need to bootstrap---and other costs associated with finetuning such as increased hardware requirements. These finetuned systems can be more efficient per test-time input, given that we currently rely on 2 open-source local calls and 1 closed-source API call for \irera{}. Our \texttt{Infer-Retrieve} program is considerably cheaper to deploy since it relies only on open-source components, while still being competitive. In the future, we plan to use an open-source LM for \texttt{Rank}, making our best program considerably cheaper to deploy as well.

\section{Conclusion}

We introduced \irera{}, a general program for extreme multi-label classification. \irera{} achieves state-of-the-art results on three benchmarks using one frozen retriever combine with two in-context learning modules. 
These findings show that the future of prompt and pipeline engineering need not be brittle. Modular programs, once optimized, can serve as highly effective general-purpose solutions.

\section*{Limitations}
The best \irera{} program currently requires one \texttt{GPT-4} call per input document, which may not feasible for all applications. In the future, we plan to explore more efficient versions of \irera{} which rely fully on low-cost open-source components.

While our optimization procedure alleviates the need for iterative prompt engineering, \irera{} does rely on an initial seed-prompt and performance may vary with spurious features of these prompts. In the future, we plan to quantify how different optimization procedures can reduce prompt brittleness, using \irera{} and our benchmark suite.

\section*{Ethics Statement}
We have applied \irera{} to the important real-world tasks of biomedical information extraction and job vacancy screening. LMs make mistakes and are biased towards certain predictions~\citep{bender2021dangers}. We advise against the deployment of \irera{} in these crucial real-world tasks without proper understanding of the risks involved and how to best measure and mitigate them.

\section*{Acknowledgements}
We are grateful to Jens-Joris Decorte and Johannes Deleu for their useful comments, and to Jens-Joris Decorte for providing us with prior statistics on the ESCO-tasks.
Karel D'Oosterlinck is funded by an FWO Fundamental Research PhD Fellowship (11632223N).
Omar Khattab is supported by the Apple Scholars in AI/ML fellowship.
This work was partially supported by IBM as a founding member of the Stanford Institute for Human-Centered Artificial Intelligence (HAI), Oracle, Virtusa, and Cigna Healthcare. 

\bibliography{anthology,custom}
\bibliographystyle{acl_natbib}

\end{document}